\documentclass[11pt]{article} % For LaTeX2e
\usepackage{rldmsubmit,palatino}
\usepackage{subcaption}
\usepackage{graphicx}
\usepackage[numbers]{natbib}
\usepackage{microtype}
\usepackage{booktabs} % for professional tables
\usepackage{amsmath, amssymb}
\usepackage{stmaryrd}
\usepackage{bm}
\usepackage{algorithm}
\usepackage{algorithmic}

\usepackage{hyperref}

\DeclareMathOperator*{\argmax}{argmax}

\usepackage{array}
\usepackage{eqparbox}

\usepackage{etoolbox}  % patch def of algorithmic environment
\makeatletter
\patchcmd{\algorithmic}{\addtolength{\ALC@tlm}{\leftmargin} }{\addtolength{\ALC@tlm}{\leftmargin}}{}{}
\makeatother

\makeatletter
\newcommand{\subalign}[1]{%
  \vcenter{%
    \Let@ \restore@math@cr \default@tag
    \baselineskip\fontdimen10 \scriptfont\tw@
    \advance\baselineskip\fontdimen12 \scriptfont\tw@
    \lineskip\thr@@\fontdimen8 \scriptfont\thr@@
    \lineskiplimit\lineskip
    \ialign{\hfil$\m@th\scriptstyle##$&$\m@th\scriptstyle{}##$\hfil\crcr
      #1\crcr
    }%
  }%
}
\makeatother

\title{Deep Reinforcement Learning with Weighted $Q$-Learning}

\author{
\hspace{-1.5cm}Andrea Cini \\
\hspace{-1.5cm}IDSIA, Universit\`a della Svizzera italiana\\
\hspace{-1.5cm}\texttt{cinia@usi.ch} \\
\And
Carlo D'Eramo \\
IAS, TU Darmstadt \\
\AND
Jan Peters \\
IAS, TU Darmstadt \\
Max Planck Institute for Intelligent Systems \\
\And
Cesare Alippi \\
IDSIA, Universit\`a della Svizzera italiana\\
Politecnico di Milano\\
}

\begin{document}

\maketitle

\begin{abstract}

Reinforcement learning algorithms based on $Q$-learning are driving Deep Reinforcement Learning~(DRL) research towards solving complex problems and achieving super-human performance on many of them. Nevertheless, $Q$-Learning is known to be positively biased since it learns by using the maximum over noisy estimates of expected values. Systematic overestimation of the action values coupled with the inherently high variance of DRL methods can lead to incrementally accumulate errors, causing learning algorithms to diverge. Ideally, we would like DRL agents to take into account their own uncertainty about the optimality of each action, and be able to exploit it to make more informed estimations of the expected return. In this regard, Weighted $Q$-Learning (WQL) effectively reduces bias and shows remarkable results in stochastic environments. WQL uses a weighted sum of the estimated action values, where the weights correspond to the probability of each action value being the maximum; however, the computation of these probabilities is only practical in the tabular setting. In this work, we provide methodological advances to benefit from the WQL properties in DRL, by using neural networks trained with Dropout as an effective approximation of deep Gaussian processes. In particular, we adopt the Concrete Dropout variant to obtain calibrated estimates of epistemic uncertainty in DRL. The estimator, then, is obtained by taking several stochastic forward passes through the action-value network and computing the weights in a Monte Carlo fashion. Such weights are Bayesian estimates of the probability of each action value corresponding to the maximum w.r.t. a posterior probability distribution estimated by Dropout. We show how our novel \emph{Deep Weighted $Q$-Learning} algorithm reduces the bias w.r.t. relevant baselines and provides empirical evidence of its advantages on representative benchmarks.

%This overoptimism can be particularly harmful in stochastic environments and when using function approximation, notably also in the case where the approximators are deep neural networks.

%Underestimation, in fact, can lead in some environments to lower performance and slower convergence rates compared to standard Q-Learning.

\end{abstract}


%\acknowledgements{We are deeply indebted to Google DeepMind and the Weinberg Institute for Cognitive Science for their generous support of RLDM2017.}  

\startmain % to start the main 1-4 pages of the submission.

\section{Introduction}
\label{s:intro}

Temporal difference~(TD) and off-policy learning are the constitutional elements of modern Reinforcement Learning~(RL). TD allows agents to bootstrap their current knowledge to learn from a new observation as soon as it is available. Off-policy learning gives the means for exploration and enables experience replay. Q-Learning~\cite{qlearning} implements both paradigms. Overestimation of the maximum action value is a well-known problem that hinders Q-Learning performance, leading to suboptimal policies and unstable learning. This overoptimism can be particularly harmful in stochastic environments and when using function approximation~\cite{thrun1993issues}, notably also in the Deep Reinforcement Learning~(DRL) settings~\cite{hasselt2016ddqn}. Among possible solutions, the Double $Q$-Learning~\cite{hasselt2010double} algorithm and its DRL variant -- Double DQN --  tackle the overestimation problem by disentangling the choice of the target action and its evaluation. The resulting estimator, while achieving superior performance in many problems, is negatively biased. In fact, overly pessimistic estimates might undervalue a good course of action and are thus problematic in their own way. In this regard, Weighted $Q$-Learning~(WQL)~\cite{deramo2016wql} is among several methods to reduce the bias and shows remarkable empirical results. WQL is based upon the Weighted Estimator~(WE), which weights each estimate of the action values, based on an estimated probability of each particular action value being the maximum. However, doing so in high-dimensional environments is not trivial due to the additional challenges imposed by using function approximation. Our objective is, then, to devise an approach to scale WQL to the DRL settings. We do so by using neural networks trained with Dropout as an effective, computationally cheap, Bayesian inference technique~\cite{gal2016dropout}. We combine, in a novel way, the dropout uncertainty estimates with the Weighted Q-Learning algorithm, extending it to the DRL settings. The proposed \textit{Deep Weighted Q-Learning} algorithm, or \textit{Weighted DQN}~(WDQN), leverages an approximated posterior distribution on Q-networks to reduce the bias of deep Q-learning. WDQN bias is neither always positive, nor negative, but depends on the state and the problem at hand. WDQN only requires minor modifications to the baseline algorithm, and its computational overhead is negligible on specialized hardware. In Section~\ref{s:bg}, we define the problem settings and discuss some related works, then, in Section~\ref{s:wdqn} we present our approach. We present empirical results in Section~\ref{s:exp} and draw our conclusions in Section~\ref{s:end}.

\section{Preliminaries}
\label{s:bg}

A Markov Decision Process~(MDP) is a tuple $\langle \mathcal{S}, \mathcal{A}, \mathcal{P}, \mathcal{R}, \gamma \rangle$ where $\mathcal{S}$ is a state space, $\mathcal{A}$ is an action space, $\mathcal{P}: \mathcal{S} \times \mathcal{A} \rightarrow \mathcal{S}$ is a Markovian transition function, $\mathcal{R}: \mathcal{S} \times \mathcal{A} \rightarrow \mathbb{R}$ is a reward function, and $\gamma \in [0,1]$ is a discount factor. A sequential decision maker ought to estimate, for each state $s$, the optimal value $Q^*(s,a)$ of each action $a$, i.e., the expected cumulative discounted reward obtained by taking action $a$ in $s$ and following the optimal policy $\pi^*$ afterwards.

\paragraph{(Deep) Q-Learning} A classical approach for solving finite MDPs is the Q-Learning algorithm, an off-policy value-based RL algorithm, based on TD. The popular Deep Q-Network algorithm (DQN)~\cite{mnih2015human} is a variant of Q-Learning designed to stabilize off-policy learning with deep neural networks in high-dimensional state spaces. The two most relevant architectural changes to standard Q-Learning introduced by DQN are the adoption of a replay memory, to learn offline from experience, and the use of a target network, to reduce correlation between the current model estimate and the bootstrapped target value. In practice, a DQN agent interacts with the environment, stores in the replay buffer performed actions and corresponding observations, and learns the Q-values online, training a neural network. The network, with parameters $\bm{\theta}$, is trained by sampling mini-batches from the memory and using a target network whose parameters $\bm{\theta}^{-}$ are updated to match those of the online model every $C$ steps. The model is trained to minimize the loss
\begin{equation}
L(\bm{\theta}) = \underset{{\langle s_i,a_i,r_i,s_i^\prime \rangle \sim m}}{\mathbb{E}}\left[\left(y_i^{DQN} - Q(s_i, a_i; \bm{\theta})\right)^2\right], \quad\quad y^{DQN}_i = r_i + \gamma \max_{a} Q(s_i^\prime, a; \bm{\theta}^{-}),
\end{equation}
where $m$ is a uniform distribution over the transitions stored in the replay buffer and $y^{DQN}_i$ is the DQN target. Among the many studied improvements and extensions of the baseline DQN algorithm, Double DQN~(DDQN)~\cite{hasselt2016ddqn} reduces the overestimation bias of DQN with a simple modification of the minimized loss. In particular, DDQN uses the target network to decouple action selection and evaluation, and estimates the target value as 
\begin{equation*}
 y^{DDQN}_i = r_i + \gamma Q(s_i^\prime,  \argmax_{a}Q\left(s_i^\prime, a; \bm{\theta}); \bm{\theta}^{-}\right).
\end{equation*}
DDQN improves on DQN, converging to a more accurate approximation of the value function, while maintaining the same model complexity and adding minimal computational overhead.

\paragraph{Estimation biases in Q-Learning} Choosing a target value for the Q-Learning update rule can be seen as an instance of the Maximum Expected Value~(MEV) estimation problem for a set of random variables, here the action values. Q-Learning uses the Maximum Estimator (ME) to estimate the maximum expected return and exploits it for policy improvement. It is well known that ME is a positively biased estimator of MEV~\cite{hasselt2013estimating}. Double Q-Learning~\cite{hasselt2010double}, on the other hand, learns two value functions in parallel and uses an update scheme based on the Double Estimator~(DE). It is shown that DE is a negatively biased estimator of MEV, which helps to avoid catastrophic overestimates of the Q-values. In practice, as also shown by~\citet{lan2020maxmin}, the overestimation bias of Q-Learning is not always harmful and may also be convenient when the action values are significantly different among each other. Conversely, the underestimation of Double Q-Learning is effective when all the action values are very similar. Unfortunately, prior knowledge about the environment is not always available, and it would be desirable to have an estimator which is not always positively or negatively polarized. Clearly, besides Double Q-Learning several other methods exist to mitigate the overestimation problem~(e.g.,~\cite{lee2013bias, anschel2017averaged, lan2020maxmin}), here we mainly focus on Weighted Q-learning and on the two most used algorithms, namely standard Q-Learning and its DE variant.

\paragraph{Weighted Q-Learning} \citet{deramo2016wql} propose the Weighted Q-Learning~(WQL) algorithm, a variant of Q-Learning based on therein introduced Weighted Estimator~(WE). WE estimates MEV as the weighted sum of the random variables sample means, weighted according to their probability of corresponding to the maximum. Intuitively, the amount of uncertainty, i.e., the entropy of the WE weights, will depend on the nature of the problem, the number of samples and the variance of the mean estimator~(critical when using function approximation). WE bias is bounded by the biases of ME and DE~\cite{deramo2016wql}. 
The target value of WQL can be computed as
\begin{equation}
\label{eq:wql}
y_t^{WQL} = r_t + \gamma \sum_{a \in \mathcal{A}} w_a^{s_{t+1}}Q(s_{t+1},a) \quad\quad w^{s}_a = P\left( a = \argmax_{a^\prime} Q(s, a^\prime)\right).
\end{equation}
where $w_a^{s_{t+1}}$ are the WE weights and correspond to the probability of each action value being the maximum. The weights of WQL are estimated in the tabular setting, assuming the sample means to be normally distributed.

\section{Deep Weighted Q-Learning}
\label{s:wdqn}
\begin{algorithm*}[ht]
   \caption{Weighted DQN}
   \label{alg:wdqn}
\begin{algorithmic}
   \STATE {\bfseries Input:} Q-network parameters $\bm{\theta}$, dropout rates $p_1, \dots, p_L$, a policy $\pi$, replay memory $\mathcal{D}$
   \STATE $\bm{\theta}^- \gets \bm{\theta}$
   %\STATE Initialize memory $\mathcal{D}$
   %\REPEAT
   \STATE Initialize memory $\mathcal{D}$.
   \FOR{step $t=0,\dots$}
   	  %\STATE Take $K$ samples $\hat{q}_k$ from $Q(s,{}\cdot{}; \bm{\theta}^- )$ by performig $K$ stochastic forward passes with MC dropout
	  \STATE Select an action $a_i$ according to some policy $\pi$ given the distribution over action-value functions $Q(s, {}\cdot{}; \bm{\theta}, \bm{\omega})$
	  \STATE Execute $a_t$ and add $\langle s_t, a_t, r_t, s_{t+i} \rangle$ to $\mathcal{D}$
	  \STATE Sample a mini-batch of transitions $\{\langle s_i,a_i,r_i,s_i^\prime \rangle, i= 1,\dots, M\}$ from $\mathcal{D}$
	  %\STATE Compute $\forall i$ the WDQN targets $y_i^{WDQN}$
	  \FOR[can be done in parallel]{$i= 1,\dots, M$}
	    \STATE Take $K$ samples from $Q(s_i,{}\cdot{}; \bm{\theta}^-, \bm{\omega})$ by performing $K$ stochastic forward passes
	    \STATE Use the samples to compute the WDQN weights~(Eq.~\ref{eq:weights}) and targets~(Eq.~\ref{eq:wdqn})
	  \ENDFOR
	  \STATE Perform an SGD step using $y^{WDQN}$ as the target value.
	  \STATE Eventually update $\bm{\theta}^-$
   %\UNTIL{$noChange$ is $true$}
   \ENDFOR
\end{algorithmic}
\end{algorithm*}

A natural way to extend the WQL algorithm to the DRL settings is to consider the uncertainty over the model parameters by using a Bayesian approach. Dropout~\cite{srivastava2014dropout} is a regularization technique used to train large neural networks by randomly dropping units during learning. In recent years, dropout has been analyzed from a Bayesian perspective~\cite{gal2016dropout}, and interpreted as a variational approximation of a posterior distribution over the parameters of the neural network. In particular, \citet{gal2016dropout} show how a neural network trained with dropout and weight decay can be seen as an approximation of a deep Gaussian process. A single stochastic forward pass through a neural network trained with Dropout can, in fact, be interpreted as taking a sample from the model's predictive distribution. This inference technique, known as \textit{Monte Carlo dropout}, can be efficiently parallelized on modern GPUs. This approach has found several applications in RL, e.g., as a practical approach to perform Thompson Sampling~\cite{gal2016dropout} and to estimate uncertainty in model-based RL~\cite{gal2015bayespilco}. Here we focus on the problem of action evaluation, and we show how to use approximate Bayesian inference to evaluate WE by introducing a novel approach to exploit uncertainty estimates in DRL. Our method is grounded in theory and simple to implement.

\paragraph{Weighted DQN} Let $Q({}\cdot{}, {}\cdot{}; \bm{\theta}, \bm{\omega})$ be a neural network with weights $\bm{\theta}$ trained with a Gaussian prior and  Dropout Variational Inference to learn the optimal action-value function of a certain MDP. We indicate with $\bm{\omega}$ the set of random variables that represents the dropout masks, with $\bm{\omega}_i$ the $i$-th realization of the random variables  and with $\bm{\Omega}$ their joint distribution:
\begin{equation}
\bm{\omega} = \{ \omega_{lk} : l= 1, \dots, L , k=1,\dots,K_{l}\}, \quad\quad \omega_{lk} \sim \text{Bernoulli}(p_l), \quad\quad \bm{\omega}_i  \sim \bm{\Omega}(p_1, \dots, p_L),
\end{equation}
where $L$ is the number of weight layers of the network and $K_l$ is the number of units in layer $l$. 
Consider a sample $q(s, a)$ of the MDP return, obtained taking action $a$ in $s$ and following the optimal policy afterwards. Following the Gaussian process interpretation of Dropout of~\citet{gal2016dropout}, we can approximate the likelihood of this \textit{observation} as a Gaussian such that
\begin{equation}
	\label{eq:gaussianapprox}
	q(s,a) \sim \mathcal{N}\left( Q(s, a; \bm{\theta}, \bm{\omega}); \tau^{-1}\right),
\end{equation}
where $\tau$ is the model precision. We can approximate the predictive mean of the process, and the expectation over the posterior distribution of the Q-value estimates, as the average of $T$ stochastic forward passes through the network:
\begin{equation}
\mathbb{E}\left[Q(s, a; \bm{\theta}, \bm{\omega})\right] \approx \hat{Q}_T(s, a; \bm{\theta}) = \frac{1}{T}\sum_{t=1}^T Q(s, a; \bm{\theta}, \bm{\omega}_t).
\end{equation}
$\hat{Q}_T(s, a; \bm{\theta})$ is the estimate of the action values associated with state $s$ and action $a$. We can estimate the probability required to calculate WE similarly, by again exploiting a Monte Carlo estimator. Given an action $a$, the probability that $a$ corresponds to the maximum expected action value can be approximated as the number of times in which, given $T$ samples, the sampled action value of $a$ is the maximum over the number of samples
\begin{equation}\label{eq:weights}
w^{s}_a(\bm{\theta}) = P\left( a = \argmax_{a^\prime} Q(s, a^\prime; \bm{\theta}, \bm{\omega})\right) \approx \frac{1}{T}\sum_{t=1}^T \left\llbracket a = \argmax_{a^\prime} Q(s, a^\prime; \bm{\theta}, \bm{\omega}_t) \right\rrbracket,
\end{equation}
where $\llbracket \dots \rrbracket$ are the Iverson brackets~($\llbracket P \rrbracket$ is $1$ if $P$ is true, $0$ otherwise). The weights can be efficiently inferred in parallel with no impact in computational time.
We can define the WE target given the target Q-network estimates by using the obtained weights as:
\begin{equation}\label{eq:wdqn}
y^{WDQN}_i = r_i + \gamma \sum_{a \in \mathcal{A}} w^{s^\prime_i}_a(\bm{\theta}^-) \hat{Q}_T(s_i^\prime, a; \bm{\theta}^-).
\end{equation}
For completeness, the complete WDQN algorithm is reported in Algorithm~\ref{alg:wdqn}. Dropout probabilities are variational parameters and influence the quality of the approximation. Ideally, they should be tuned to maximize the log-likelihood of the observations using a validation method. This is clearly not possible in RL where the available samples and the underlying distribution generating them, change as the policy improves. In fact, using dropout with a fixed probability might lead to poor uncertainty estimates~\cite{osband2016boot, osband2018deeprandomized, gal2017concrete}. Concrete Dropout~\cite{gal2017concrete} mitigates this problem by using a differentiable continuous relaxation of the Bernoulli distribution and learning the dropout rate from data. In practice, then, we substitute the standard Dropout with its Concrete variant in our WDQN implementation.

\section{Experiments}
\label{s:exp}

\begin{figure*}[ht]
    \centering
    
    \begin{subfigure}[b]{0.49\columnwidth}
         \centering
         \includegraphics[width=\textwidth]{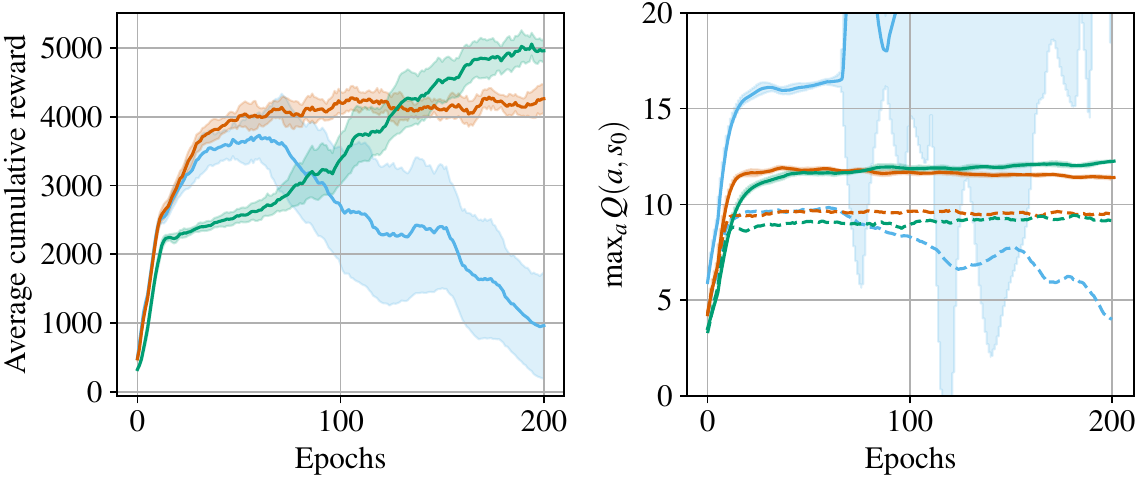}
         \caption{Asterix.}
         \label{fig:y equals x}
     \end{subfigure}
     \begin{subfigure}[b]{0.49\columnwidth}
         \centering
         \includegraphics[width=\textwidth]{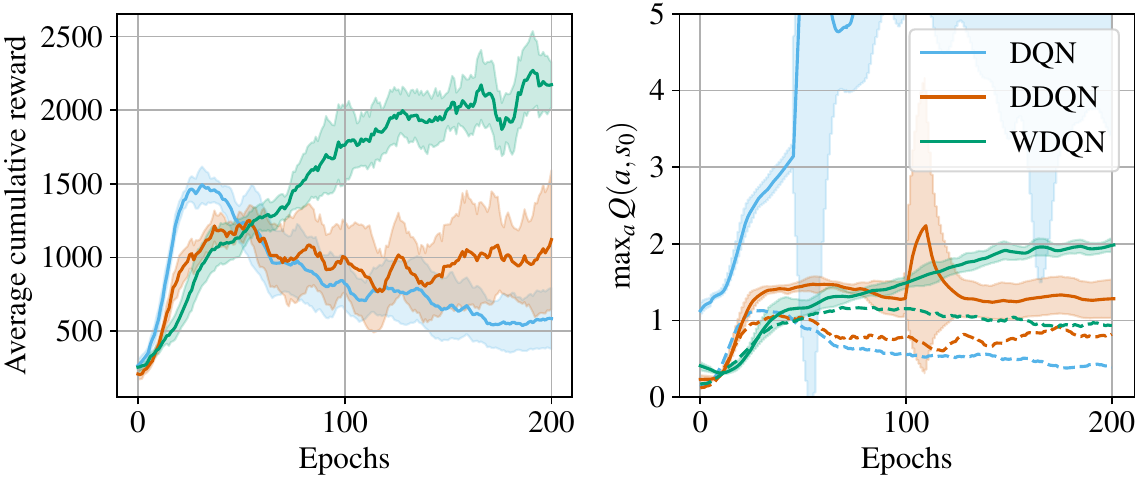}
         \caption{Wizard Of Wor.}
         \label{fig:three sin x}
     \end{subfigure}
    \caption{Learning curves on Atari games. The curves on the left of each subfigure show the evaluation scores. The curves on the right, conversely, report the estimate of the expected return w.r.t. the starting screen of the game; the dashed curves indicate the real discounted return for each agent. Each epoch corresponds to 1M frames. The results shown here are the average of $10$ independent runs and the shaded areas represent $95\%$ confidence intervals. The curves are smoothed using a moving average of 10 epochs to improve readability.}
    \label{fig:atari}
\end{figure*}

In this section, we compare WDQN against the standard DQN algorithm and its Double DQN variant in the Arcade Learning Environment~(ALE)~\cite{bellemare13arcade}. In particular, we choose environments where DQN is known to overestimate action values and exhibit unstable learning~\cite{hasselt2016ddqn}. We use the same neural network and hyperparameters of~\citet{mnih2015human}, for WDQN we use Concrete Dropout only in the fully connected layer after the convolutional block. We found WDQN to be robust to the number of dropout samples used to compute the WE~(e.g., $T \geq 30$), while being more sensitive to the Concrete Dropout regularization coefficient. For more in-depth results and discussion, we refer to our journal paper~\cite{deramo2021gaussian}. 

Figure~\ref{fig:atari} shows the result of the comparison in terms of the average reward and prediction accuracy of WDQN against DQN and DDQN. Here, WDQN stabilizes learning in both scenarios, but it also outperforms DDQN. Conversely, DDQN manages to stabilize the learning performance in Asterix, but, in our setting, is not able to yield satisfactory performance in Wizard of Wor differently from what observed by~\citet{hasselt2016ddqn}, probably due to minor implementation differences. However, it is worth mentioning that DDQN would most likely improve in this scenario by reducing the frequency of updating the target network's parameters. WDQN, on the other hand, manages to control the bias and the learning instability in both cases, achieving a comparable performance and prediction accuracy w.r.t. DDQN in Asterix and outperforming the other two baselines on Wizard of Wor. Again, we refer to the journal paper for additional results.

\section{Conclusion and future works}
\label{s:end}

We present WDQN, a new value-based Deep Reinforcement Learning algorithm that extends the Weighted Q-Learning algorithm to work in environments with a high-dimensional state representation. WDQN is a principled and robust method to exploit uncertainty in DRL to accurately estimate the maximum action value in a given state. Our results corroborate the findings of previous works~\cite{gal2016dropout, gal2017concrete}, confirming that dropout can be used successfully for approximate Bayesian inference in DRL. Future works may explore the combination of WDQN with other orthogonal DQN extensions and may attempt to adapt the WDQN approach to other techniques modeling uncertainty in deep neural networks.

\bibliographystyle{plainnat}
\bibliography{biblio}

\end{document}